\documentclass[10pt,twocolumn,letterpaper]{article}

\usepackage[pagenumbers]{cvpr}       
\usepackage{booktabs}
\usepackage{multirow}
\usepackage{graphicx}
\usepackage{tabularx}
\usepackage{array}
\usepackage{microtype}
\usepackage{algorithm}
\usepackage{algpseudocode}

\definecolor{cvprblue}{rgb}{0.21,0.49,0.74}
\usepackage[pagebackref,breaklinks,colorlinks,allcolors=cvprblue]{hyperref}

\def\confName{CVPR}
\def\confYear{2027} 

\title{Learning When Skills Work: Boundary-Aware Composition for LLM Agents}
\author{
Dingyu Zhang$^{1,*}$ \quad
Jiaxin Qi$^{2,*}$ \quad
Yao Tong$^{1}$\\
Yan Cui$^{1}$ \quad
Yuhua Zheng$^{1}$ \quad
Jianqiang Huang$^{2}$\\[4pt]
{\fontsize{11}{13}\selectfont
$^{1}$Hangzhou Institute for Advanced Study,
University of Chinese Academy of Sciences, Hangzhou, China}\\
{\fontsize{11}{13}\selectfont
$^{2}$Computer Network Information Center,
Chinese Academy of Sciences, Beijing, China}\\
{\normalfont\rmfamily\large
zhangdingyu26@mails.ucas.ac.cn,
jxqi@cnic.cn,
tongyao261@mails.ucas.ac.cn,}\\
{\normalfont\rmfamily\large
cuiyan.ch@gmail.com,
zhengyuhua@ucas.ac.cn,
jqhuang@cnic.cn
}\\[2pt]
{\small $^{*}$Equal contribution}
}

\begin{document}
\maketitle
\begin{abstract}
Skill optimization enhances agent capabilities by iteratively refining reusable skill instructions from execution feedback. Guided by distinct strategies and priorities, different optimization methods yield specialized skills that each excel at specific behaviors. Skill composition has therefore been explored to consolidate these skills into a unified one. However, existing composition approaches suffer from three critical problems: instruction conflicts, behavioral granularity mismatches, and context-insensitive behavior transfer. To resolve these problems, we propose History-Based Behavior Transfer (HBT), a framework for conditional skill composition. First, HBT selects a main skill from a set of candidate optimized skills to establish a coherent structural foundation. Second, it isolates behavioral differences from the remaining optimized skills and links them to task-level gains and regressions in their optimization histories. Finally, by contrasting these beneficial and detrimental outcomes, HBT infers explicit applicability conditions for each candidate behavior and integrates the resulting condition-behavior pairs into the main skill for selective activation. Extensive evaluations across diverse benchmarks and skill optimizers demonstrate that HBT consistently outperforms both the best single optimized skills and existing composition baselines. Further ablations on historical evidence and conditional gating confirm the effectiveness of HBT in achieving reliable skill composition.

\end{abstract}

\section{Introduction}
\label{sec:intro}

\begin{figure}[t]
    \centering
    \includegraphics[width=\columnwidth]{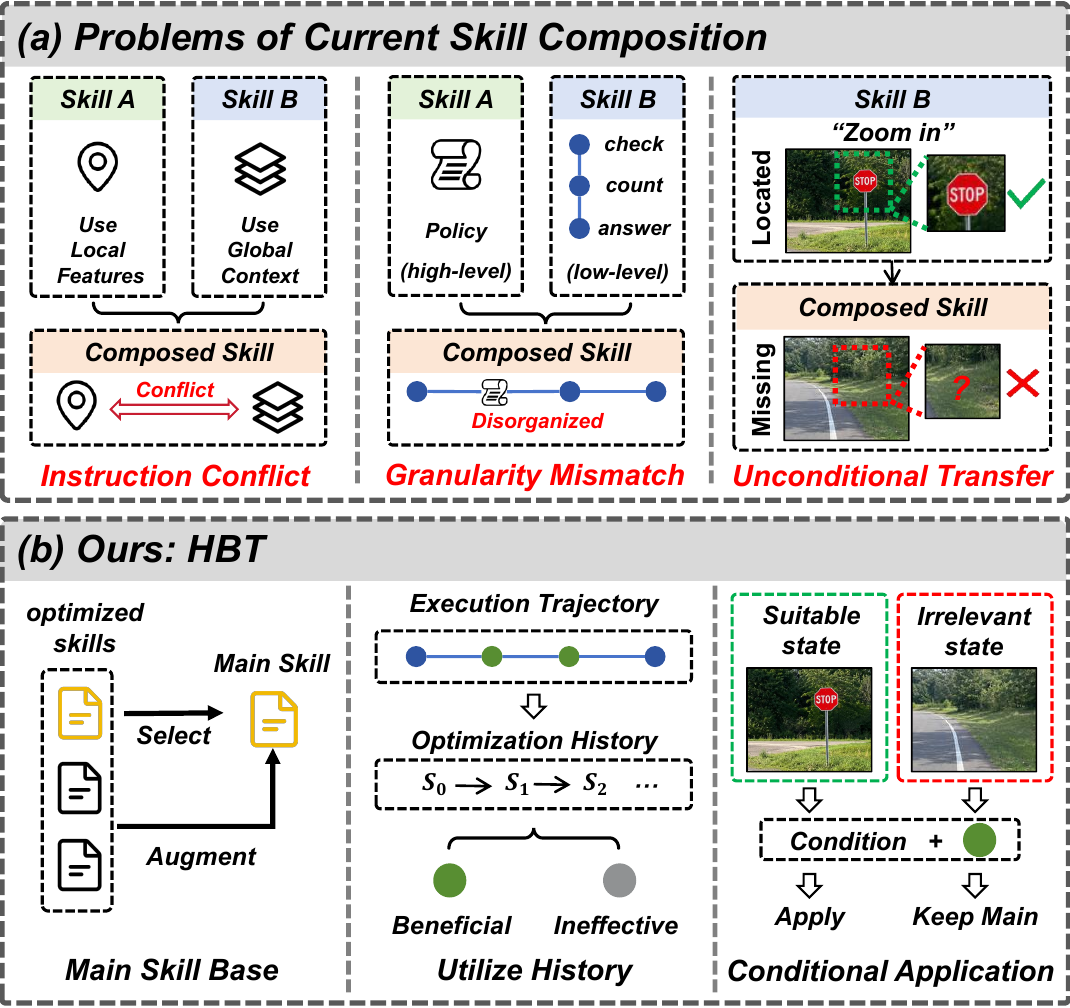}
    \caption{Overview of current skill composition problems and our HBT framework. (a) Directly merging multiple optimized skills induces instruction conflicts, behavioral granularity mismatches, and context-insensitive behavior transfer. (b) HBT builds upon a selected main skill, traces execution trajectories across optimization histories to separate beneficial behaviors from ineffective ones, and infers applicability conditions to activate transferred behaviors only in suitable states.}
    \label{fig:teaser}
\end{figure}

Skill optimization aims to improve LLM agents by iteratively refining reusable skills from execution experience~\cite{yao2022react,shinn2023reflexion,wang2023voyager,ni2026trace2skill}. In each iteration, an agent executes tasks with the current skill while an optimizer updates the instructions through discrete textual revisions, recording intermediate versions, edits, and task-level outcomes as an optimization history. As a single update can fix specific failures while inducing regressions on other tasks~\cite{nie2026skillevoreg}, different optimization methods adopt distinct revision and selection mechanisms to govern which changes are retained. For instance, SkillOpt applies small, conservative edits gated by overall validation accuracy~\cite{yang2026skillopt}, yielding broadly reliable, general-purpose strategies; by contrast, SkillGrad tracks recurring failures across iterations to make targeted fixes~\cite{wang2026skillgrad}, uncovering specialized heuristics for challenging edge cases. Even when starting from the same initial skill, these contrasting mechanisms yield optimized skills with distinct behavioral profiles. Consequently, even the highest-scoring individual skill may still miss effective behaviors captured by other optimized skills.

To consolidate these specialized behaviors within a single agent, skill composition aims to integrate multiple optimized skills into a unified one. Existing approaches either directly merge and reconcile skill instructions~\cite{zhang2026skillcomposer,li2026skillaligner} or aggregate reusable insights and behavioral patches distilled from execution trajectories~\cite{zhao2024expel,chen2026skillcat}. As illustrated in Fig.~\ref{fig:teaser}(a), however, existing composition approaches suffer from three critical problems: (1) instruction conflicts, where skills prescribe competing decision rules~\cite{yan2026diagnosing,zhang2026many}; for instance, one skill prioritizes fine-grained local features while another emphasizes global scene context, causing the agent to oscillate between contradictory actions and make erratic decisions when local and global cues disagree; (2) behavioral granularity mismatches, where abstract, high-level policies are combined with concrete, low-level action sequences (e.g., a sequential check, count, and answer procedure) without structural coordination, disrupting the execution flow and causing disorganized behavior~\cite{wen2024benchmarking}; and (3) context-insensitive behavior transfer, where a specialized behavior from one skill is transferred into the composed skill without the prerequisite state under which it succeeds. For example, one skill may include a zooming-in action that inspects fine details once a target is roughly localized, solving hard cases where another skill fails; yet transferring this action unconditionally causes the composed skill to zoom in even when the target is not yet localized, focusing on irrelevant background and inducing regressions on previously solved tasks.

Motivated by these problems, we propose \textbf{History-Based Behavior Transfer (HBT)}, a framework for conditional skill composition. First, HBT selects a single main skill from a set of candidate optimized skills to establish a coherent structural foundation, preserving one primary decision hierarchy while incorporating new capabilities strictly as modular supplements. Grounding composition in a unified backbone reduces the instruction conflicts that arise when competing rule sets are merged together. Second, it isolates behavioral differences from the remaining optimized skills and links them to task-level gains and regressions across historical revisions in their optimization histories. By aligning the relevant decision stages around these revisions, HBT grounds each difference in a concrete, well-scoped behavioral unit with a clear execution role, mitigating behavioral granularity mismatches. Finally, HBT contrasts pre-decision execution states across these beneficial and detrimental outcomes to infer explicit applicability conditions for each candidate behavior. The resulting condition-behavior pairs are then integrated into the main skill, enabling the agent to selectively activate transferred behaviors when matching conditions arise and thereby mitigating context-insensitive behavior transfer.

Through extensive experiments across five benchmarks and four skill optimizers, we evaluate HBT against both individual optimized skills and existing skill composition baselines. Across multiple LLM backbones, HBT consistently achieves the highest overall performance, outperforming the strongest single optimized skills and existing composition baselines by 2.70 to 5.80 percentage points on interactive agent benchmarks, while surpassing the best single skill by 2.79 percentage points to reach 84.16\% average accuracy on visual reasoning. Furthermore, comprehensive ablation studies confirm the necessity and effectiveness of each core module in HBT, including behavioral difference extraction, behavioral boundary localization, and contrastive condition inference. Detailed behavioral analyses further verify that the inferred condition-behavior pairs preserve performance on previously solved tasks while selectively activating specialized behaviors in matching contexts, demonstrating the soundness and reliability of our design.

In summary, our main contributions are threefold:
\begin{itemize}
    \item We identify three critical problems in existing skill composition approaches: instruction conflicts, behavioral granularity mismatches, and context-insensitive behavior transfer across different execution states.

    \item We propose HBT, a history-guided framework for conditional skill composition that integrates behavioral difference extraction, behavioral boundary localization, and contrastive condition inference into a unified pipeline built around a main skill backbone.
    
    \item Extensive experiments across five benchmarks and four skill optimizers demonstrate that HBT consistently outperforms individual skills and composition baselines, while ablation studies further validate the contribution of each architectural component to the overall performance.
\end{itemize}

\section{Related Work}
\label{sec:related_work}
\subsection{Agent Skill Optimization}

Research on agent skill optimization has progressed from extracting reusable procedures from execution experience to iteratively refining skills through task feedback~\cite{fernando2023promptbreeder,liu2026skillforge}. Existing methods explore mechanisms for generating and gating updates, including feedback-driven reflection, iterative candidate search, and evaluation-guided selection~\cite{pryzant2023automatic,liu2026skillrevise,chen2026vce}, alongside strategies to regulate revisions and mitigate overfitting across optimization rounds~\cite{moll2026grasp,lin2026rethinking}. Beyond producing a final skill, multi-round optimization accumulates intermediate feedback, edits, and evaluation outcomes, which recent work has begun leveraging to guide subsequent updates and analyze optimization stability~\cite{li2026skilladam}. Because discrete textual revisions often fix targeted failures while inducing regressions elsewhere, optimized skills frequently acquire specialized behaviors that are effective only under specific states rather than globally applicable. However, leveraging optimization histories to assess the transferability of such behaviors across independently optimized skills and infer their applicability conditions remains underexplored. To bridge this gap, HBT traces historical gains and regressions to distill explicit conditions for skill composition.

\begin{figure*}[t]
    \centering
    \includegraphics[width=\textwidth]{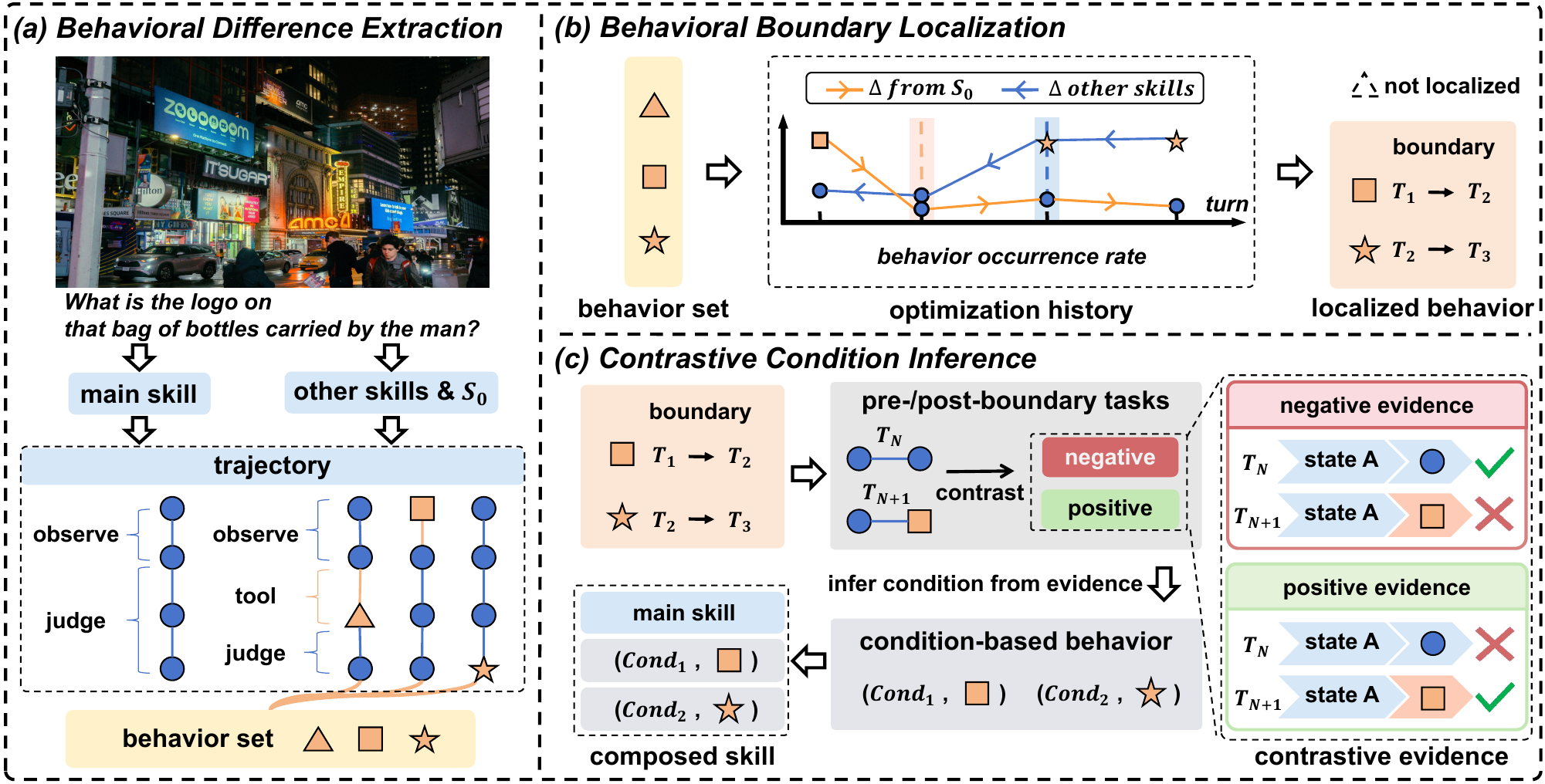}
    \caption{Overview of HBT. (a) Behavioral Difference Extraction compares execution trajectories of the main skill against the initial skill $S_0$ and other optimized skills to build a candidate behavior set. (b) Behavioral Boundary Localization tracks behavior occurrence rates across optimization histories to identify the revision boundaries with the sharpest behavioral shifts. (c) Contrastive Condition Inference contrasts positive evidence and negative evidence across boundary revisions to infer applicability conditions, appending the resulting condition-behavior pairs to the main skill to form the composed skill.}
    \label{fig:method}
\end{figure*}

\subsection{Skill Composition}

Offline skill composition consolidates knowledge from multiple skills or experiences into a unified artifact. Instruction-level approaches reconcile conflicts across skill descriptions, adapt retrieved skills prior to execution~\cite{li2026skillaligner}, or learn merging operators~\cite{zhang2026skillcomposer}, yet static text integration cannot reveal which behavioral differences drive task gains versus regressions. Trajectory-based methods distill reusable lessons from execution rollouts~\cite{zhao2024expel,ni2026trace2skill}, but post-hoc distillation discards iterative revision provenance across optimization rounds. Orthogonal efforts explore context-dependent skill selection~\cite{ahn2022can,sun2023adaplanner} and applicability boundaries from interaction memory~\cite{tang2026memory}, focusing on runtime routing or online memory governance rather than cross-skill behavior transfer. Consequently, leveraging optimization histories across independently optimized skills to evaluate behavioral differences and infer pre-decision applicability conditions remains unexplored. To address this, HBT anchors behavioral supplements onto a main skill backbone with pre-decision conditions inferred from historical outcome shifts.

\section{Method}
\label{sec:method}

Figure~\ref{fig:method} illustrates the overall framework of HBT, consisting of behavioral difference extraction, behavioral boundary localization, and contrastive condition inference.

\subsection{Problem Formulation}
We model an LLM agent as a skill-conditioned policy $\pi(\cdot \mid x, S)$ that solves a task $x \in \mathcal{D}$ by generating an execution trajectory $\tau \sim \pi(\cdot \mid x, S)$ and receiving a binary outcome $r(x, \tau) \in \{0, 1\}$ indicating task failure or success. Starting from an initial skill $S_0$, skill optimization iteratively refines the skill instructions to maximize the pass rate on $\mathcal{D}$. When $K$ independent optimizers $\{\mathcal{O}_i\}_{i=1}^K$ are applied to $(S_0, \mathcal{D})$, each produces an optimized skill together with its optimization history:
\begin{equation}
(S_i, H_i) = \mathcal{O}_i(S_0, \mathcal{D}), \quad i = 1, \ldots, K,
\end{equation}
where $H_i$ records the intermediate skill versions, revision steps, and per-task execution trajectories and outcomes accumulated across optimization iterations.

Given $S_0$ and the optimization outputs $\{(S_i, H_i)\}_{i=1}^K$, offline skill composition aims to synthesize a unified skill $S_c$ from these skills and their optimization histories to maximize expected pass rate:
\begin{equation}
\max_{S_c} \; \mathbb{E}_{x \in \mathcal{D},\, \tau \sim \pi(\cdot \mid x, S_c)} \bigl[r(x, \tau)\bigr].
\end{equation}

\subsection{Behavioral Difference Extraction}
\label{sec:behavioral-discovery}

We first designate the skill in $\{S_i\}_{i=1}^K$ achieving the highest pass rate on $\mathcal{D}$ under repeated evaluation as the main skill $S_m$, which serves as the foundation for composition. We then compare the execution trajectories of $S_m$ with those of $S_0$ and the remaining optimized skills on tasks in $\mathcal{D}$. Let $\tau_i(x)$ denote the execution trajectory of $S_i$ on task $x$ ($i = 0, 1, \ldots, K$), and let $\tau_m(x)$ denote that of $S_m$. For each task on which a non-main skill $S_i \neq S_m$ succeeds ($r(x, \tau_i(x)) = 1$), regardless of whether $S_m$ succeeds or fails, an LLM compares $\tau_i(x)$ against $\tau_m(x)$ based on executed actions and resulting environmental feedback to extract decision behaviors exhibited by $S_i$ but absent from $\tau_m(x)$.

We then use an LLM to consolidate semantically similar behavioral differences across tasks into a set of candidate behaviors $\mathcal{B}$~\cite{zheng2023judging}. Each candidate behavior $b \in \mathcal{B}$ retains associations with its originating skill(s) for subsequent historical boundary localization.

\subsection{Behavioral Boundary Localization}

For each candidate behavior $b \in \mathcal{B}$, we trace its evolution across the optimization history $H_i$ of its originating skill $S_i$, or across the optimization history of $S_m$ when $b$ originates from $S_0$. Within an optimization history of $T$ rounds, let $\tau^{(t)}(x)$ denote the execution trajectory of the skill version at round $t$ on task $x \in \mathcal{D}$. We measure the occurrence rate of $b$ at round $t$ as
\begin{equation}
\rho_t(b) = \frac{1}{|\mathcal{D}|} \sum_{x \in \mathcal{D}} \mathbb{I}[b \in \tau^{(t)}(x)],
\end{equation}
where $\mathbb{I}[b \in \tau^{(t)}(x)]$ indicates whether $b$ occurs in $\tau^{(t)}(x)$, determined by matching its description against the executed actions and resulting environmental feedback. We localize the behavioral boundary at the transition $(n, n+1)$ exhibiting the largest shift in occurrence rate:
\begin{equation}
n = \arg\max_{t \in \{0, \ldots, T-1\}} \left|\rho_{t+1}(b) - \rho_t(b)\right|.
\end{equation}
Accordingly, we denote the two adjacent skill versions immediately before and after the identified boundary, at rounds $n$ and $n+1$, by $S^{(n)}$ and $S^{(n+1)}$, respectively.

\subsection{Contrastive Condition Inference}

With the behavioral boundary $(n, n+1)$ localized for a candidate behavior $b$, we examine the trade-off tasks in $\mathcal{D}$ whose outcomes flip between $S^{(n)}$ and $S^{(n+1)}$, namely tasks satisfying $r(x, \tau^{(n)}(x)) \neq r(x, \tau^{(n+1)}(x))$. For each such task $x$, the execution trajectories of $S^{(n)}$ and $S^{(n+1)}$ yield opposite outcomes, with the successful trajectory serving as a in-task reference for subgoal completion. We carefully align the decision stages relevant to $b$ across both trajectories and compare their corresponding executed actions and resulting observations against this reference to assess whether the observed use of $b$ is associated with subgoal progress or regression, discarding cases with ambiguous local evidence.

From the retained cases, we group the pre-decision states into those associated with beneficial versus detrimental outcomes. An LLM then contrasts these two groups to infer an applicability condition $\mathrm{cond}_b$ grounded solely in state features observable prior to the decision.

Finally, an LLM verifies each candidate pair $(\mathrm{cond}_b, b)$ for consistency with its supporting historical evidence and observability at decision time, filtering out unsupported or unreliable rules~\cite{dhuliawala2024chain}. The remaining $M$ validated pairs $\{(\mathrm{cond}_j, b_j)\}_{j=1}^M$ are appended to $S_m$ to construct the composed skill $S_c = S_m \oplus \{(\mathrm{cond}_j, b_j)\}_{j=1}^M$. At execution time, the agent activates $b_j$ only when $\mathrm{cond}_j$ is satisfied, and otherwise follows the original instructions of $S_m$. The overall procedure is summarized in Algorithm~\ref{alg:hbt}.

\begin{algorithm}[t]
\caption{HBT: History-Based Behavior Transfer}
\label{alg:hbt}
\small
\renewcommand{\algorithmicrequire}{\textbf{Inputs}}
\renewcommand{\algorithmicensure}{\textbf{Output}}
\algrenewcommand{\algorithmiccomment}[1]{\hfill\textcolor{gray}{\textit{// #1}}}
\begin{algorithmic}[1]
\Require $S_0, \{S_i\}, \{H_i\}, \mathcal{D}$
\Ensure $S_c$
\State Evaluate $S_0, \{S_i\}$ on $\mathcal{D}$ to obtain execution trajectories $\mathcal{T}_0, \{\mathcal{T}_i\}$ \Comment{Stage 1}
\State Select the highest-pass-rate main skill $S_m$ and extract candidate behaviors $\mathcal{B}$
\State Initialize retained pairs $\mathcal{P} \gets \emptyset$
\For{each $b \in \mathcal{B}$}
    \State Select the relevant history $H_b$ for $b$ \Comment{Stage 2}
    \State Locate boundary $n = \arg\max_t |\rho_{t+1}(b)-\rho_t(b)|$ over adjacent rounds in $H_b$
    \State Identify trade-off tasks $\mathcal{E}$ with flipped outcomes between $S^{(n)}$ and $S^{(n+1)}$ in $H_b$ \Comment{Stage 3}
    \State Align decision stages on $\mathcal{E}$ using successful trajectories as references
    \State Collect positive and negative pre-decision states $\mathcal{Z}_b^+, \mathcal{Z}_b^-$
    \If{$\mathcal{Z}_b^+ \neq \emptyset$}
        \State Infer $\mathrm{cond}_b$ from $\mathcal{Z}_b^+, \mathcal{Z}_b^-$
        \State Validate $(\mathrm{cond}_b,b)$ and add it to $\mathcal{P}$ if valid
    \EndIf
\EndFor
\State $S_c \gets S_m \oplus \mathcal{P}$
\State \Return $S_c$
\end{algorithmic}
\end{algorithm}

\section{Experiments}
\label{sec:experiments}

\newcolumntype{C}[1]{>{\centering\arraybackslash}p{#1}}

\begin{table*}[t]
    \centering
    \small
    \setlength{\tabcolsep}{2pt}
    \renewcommand{\arraystretch}{1.08}

    \noindent
    \begin{minipage}[t]{0.49\textwidth}
        \vspace{0pt}
        \centering
        {\normalsize\bfseries GLM-5.3-Flash}\\[3pt]
        \begin{tabular*}{\linewidth}
        {@{\extracolsep{\fill}}p{1.42cm}*{6}{C{1.02cm}}@{}}
            \toprule
            \raisebox{-5pt}[0pt][0pt]{%
                \multirow[c]{2}{1.42cm}{\makebox[1.42cm][c]{Method}}
            } &
            \multicolumn{5}{c}{Benchmark Success Rate (\%)} &
            \raisebox{-5pt}[0pt][0pt]{%
                \multirow[c]{2}{1.02cm}{\makebox[1.02cm][c]{Avg.}}
            } \\
            \cmidrule(lr){2-6}
            & \shortstack{Science\\World}
            & \shortstack{Spread-\\sheet}
            & \shortstack{Deep\\Planning}
            & \shortstack{App\\World}
            & \shortstack{$\tau^3$-\\Bench}
            & \\
            \midrule
            Init Skill    & 35.00 & 65.00 & 35.00 & 59.52 & 79.85 & 54.87 \\
            SkillOpt      & 19.17 & \textbf{73.57} & 45.00 & 70.83 & 82.09 & 58.13 \\
            SkillHone     & 30.00 & 67.14 & 48.75 & 68.45 & 79.85 & 58.84 \\
            EvoSkill      & 40.00 & 68.21 & 40.00 & 61.90 & 82.84 & 58.59 \\
            SkillEvolver  & 38.33 & 67.86 & 41.25 & 65.48 & 82.09 & 59.00 \\
            \cmidrule(lr){1-7}
            Trace2Skill   & 50.00 & 68.21 & 51.25 & 60.71 & 82.09 & 62.45 \\
            RCR           & 37.50 & 70.71 & 45.00 & 70.24 & 79.85 & 60.66 \\
            HBT & \textbf{51.67} & 73.21 & \textbf{53.75} & \textbf{76.79} & \textbf{85.82} & \textbf{68.25} \\
            \bottomrule
        \end{tabular*}
    \end{minipage}
    \hfill
    \raisebox{1.9pt}[0pt][0pt]{%
    \begin{minipage}[t]{0.49\textwidth}
        \vspace{0pt}
        \centering
        {\normalsize\bfseries DeepSeek-V4.1-Flash}\\[3pt]
        \begin{tabular*}{\linewidth}
        {@{\extracolsep{\fill}}p{1.42cm}*{6}{C{1.02cm}}@{}}
            \toprule
            \raisebox{-5pt}[0pt][0pt]{%
                \multirow[c]{2}{1.42cm}{\makebox[1.42cm][c]{Method}}
            } &
            \multicolumn{5}{c}{Benchmark Success Rate (\%)} &
            \raisebox{-5pt}[0pt][0pt]{%
                \multirow[c]{2}{1.02cm}{\makebox[1.02cm][c]{Avg.}}
            } \\
            \cmidrule(lr){2-6}
            & \shortstack{Science\\World}
            & \shortstack{Spread-\\sheet}
            & \shortstack{Deep\\Planning}
            & \shortstack{App\\World}
            & \shortstack{$\tau^3$-\\Bench}
            & \\
            \midrule
            Init Skill    & 26.67 & 77.86 & 48.75 & 50.00 & 84.33 & 57.52 \\
            SkillOpt      & 50.00 & 77.14 & 55.00 & 65.48 & 82.84 & 66.09 \\
            SkillHone     & 48.33 & 76.07 & \textbf{63.75} & 83.93 & 86.57 & 71.73 \\
            EvoSkill      & 55.83 & 76.07 & 51.25 & 63.69 & 85.82 & 66.53 \\
            SkillEvolver  & 47.50 & 78.93 & \textbf{63.75} & 92.86 & \textbf{88.81} & 74.37 \\
            \cmidrule(lr){1-7}
            Trace2Skill   & 57.50 & 76.79 & 58.75 & 89.29 & 85.82 & 73.63 \\
            RCR   & 55.83 & 76.79 & 62.50 & 93.45 & 85.07 & 74.73 \\
            HBT & \textbf{59.17} & \textbf{82.14} & \textbf{63.75} & \textbf{94.05} & 88.06 & \textbf{77.43} \\
            \bottomrule
        \end{tabular*}
    \end{minipage}%
    }

    \vspace{10pt}

    \noindent
    \begin{minipage}[t]{0.49\textwidth}
        \vspace{0pt}
        \centering
        {\normalsize\bfseries GPT-6-Luna}\\[3pt]
        \begin{tabular*}{\linewidth}
        {@{\extracolsep{\fill}}p{1.42cm}*{6}{C{1.02cm}}@{}}
            \toprule
            \raisebox{-5pt}[0pt][0pt]{%
                \multirow[c]{2}{1.42cm}{\makebox[1.42cm][c]{Method}}
            } &
            \multicolumn{5}{c}{Benchmark Success Rate (\%)} &
            \raisebox{-5pt}[0pt][0pt]{%
                \multirow[c]{2}{1.02cm}{\makebox[1.02cm][c]{Avg.}}
            } \\
            \cmidrule(lr){2-6}
            & \shortstack{Science\\World}
            & \shortstack{Spread-\\sheet}
            & \shortstack{Deep\\Planning}
            & \shortstack{App\\World}
            & \shortstack{$\tau^3$-\\Bench}
            & \\
            \midrule
            Init Skill    & 39.17 & 58.57 & 37.50 & 67.26 & 58.96 & 52.29 \\
            SkillOpt      & 44.17 & 58.57 & 37.50 & 69.64 & 58.96 & 53.77 \\
            SkillHone     & 39.17 & 55.71 & 46.25 & 62.50 & 59.70 & 52.67 \\
            EvoSkill      & 39.17 & 50.71 & 41.25 & 67.26 & 58.96 & 51.47 \\
            SkillEvolver  & 48.33 & 58.93 & 38.75 & 70.24 & 55.22 & 54.29 \\
            \cmidrule(lr){1-7}
            Trace2Skill   & 26.67 & 56.79 & 47.50 & 67.86 & \textbf{62.69} & 52.30 \\
            RCR           & 44.17 & 57.14 & 53.75 & 66.25 & 57.46 & 55.75 \\
            HBT & \textbf{50.00} & \textbf{59.64} & \textbf{58.75} & \textbf{74.40} & 59.70 & \textbf{60.50} \\
            \bottomrule
        \end{tabular*}
    \end{minipage}
    \hfill
    \raisebox{1.9pt}[0pt][0pt]{%
    \begin{minipage}[t]{0.49\textwidth}
        \vspace{0pt}
        \centering
        {\normalsize\bfseries Qwen3.8-Max}\\[3pt]
        \begin{tabular*}{\linewidth}
        {@{\extracolsep{\fill}}p{1.42cm}*{6}{C{1.02cm}}@{}}
            \toprule
            \raisebox{-5pt}[0pt][0pt]{%
                \multirow[c]{2}{1.42cm}{\makebox[1.42cm][c]{Method}}
            } &
            \multicolumn{5}{c}{Benchmark Success Rate (\%)} &
            \raisebox{-5pt}[0pt][0pt]{%
                \multirow[c]{2}{1.02cm}{\makebox[1.02cm][c]{Avg.}}
            } \\
            \cmidrule(lr){2-6}
            & \shortstack{Science\\World}
            & \shortstack{Spread-\\sheet}
            & \shortstack{Deep\\Planning}
            & \shortstack{App\\World}
            & \shortstack{$\tau^3$-\\Bench}
            & \\
            \midrule
            Init Skill    & 30.83 & 86.79 & 63.75 & 79.17 & 81.34 & 68.38 \\
            SkillOpt      & 61.67 & 87.14 & 63.75 & 80.95 & 86.57 & 76.02 \\
            SkillHone     & \textbf{100.00} & 85.00 & 71.25 & 79.17 & 85.82 & 84.25 \\
            EvoSkill      & 74.17 & 85.71 & 66.25 & 86.31 & 85.07 & 79.50 \\
            SkillEvolver  & 84.17 & 88.21 & 53.75 & \textbf{94.05} & 86.57 & 81.35 \\
            \cmidrule(lr){1-7}
            Trace2Skill   & \textbf{100.00} & 88.93 & 60.00 & 92.26 & 88.06 & 85.85 \\
            RCR           & 95.83 & 87.14 & 71.25 & 91.67 & 82.09 & 85.60 \\
            HBT & \textbf{100.00} & \textbf{91.07} & \textbf{76.25} & \textbf{94.05} & \textbf{91.79} & \textbf{90.63} \\
            \bottomrule
        \end{tabular*}
    \end{minipage}%
    }

    \caption{Main results (task success rate, \%) across five benchmarks and four base models. For each backbone, we compare the initial skill and four individual skill optimizers against skill composition baselines and HBT. Avg. denotes the arithmetic mean across the five benchmarks. Best results in each benchmark column are bolded.}
    \label{tab:main_results}
\end{table*}
\subsection{Experimental Setup}

\noindent\textbf{Benchmarks.} We evaluate our method on five benchmarks spanning diverse agent tasks: \textbf{ScienceWorld} (Genetics)~\cite{wang2022scienceworld}, \textbf{Spread\-sheet\-Bench} (Verified)~\cite{ma2024spreadsheetbench}, \textbf{Deep\-Plan\-ning} (Shopping)~\cite{zhang2026deepplanning}, \textbf{AppWorld} (official Test-Normal split)~\cite{trivedi2024appworld}, and $\boldsymbol{\tau^3}$\textbf{-Bench} (Retail and Airline)~\cite{yao2024tau}. All compared methods are evaluated on the same task sets under consistent evaluation protocols and settings.

\noindent\textbf{Skill Optimizers.} We employ four representative skill optimization methods to generate optimized skills and collect their corresponding optimization histories: SkillOpt~\cite{yang2026skillopt}, which performs feedback-driven skill refinement through bounded textual edits and validation-based selection; SkillHone~\cite{li2026skillhone}, which leverages persistent decision histories to guide iterative skill refinement; EvoSkill~\cite{alzubi2026evoskill}, which discovers and refines skills through failure analysis and evolutionary selection; and SkillEvolver~\cite{zhang2026skillevolver}, which employs a meta-skill to iteratively author, deploy, and refine domain-specific skills based on execution feedback.

\noindent\textbf{Baselines.} We compare our method with the initial skill $S_0$, the four independently optimized skills, Trace2Skill~\cite{ni2026trace2skill}, and a three-call LLM-based Rewrite--Critique--Revise (RCR) baseline. For Trace2Skill, we provide the training-set execution trajectories generated by the four optimized skills as input and use its output as the composed skill. RCR takes the four optimized skills as input and applies a three-step Rewrite--Critique--Revise process to produce a unified skill. The compared methods leverage different types and amounts of execution evidence, resulting in different computational requirements. We therefore focus our evaluation on the effectiveness of the resulting skill compositions.

\subsection{Implementation Details}
Since the benchmarks do not provide predefined initial skills, we generate a benchmark-specific initial skill $S_0$ for each LLM backbone using the benchmark and training-task descriptions. The generation excludes reference solutions, execution trajectories, and evaluation feedback, and aims to produce task-solving guidance. Each benchmark uses fixed splits of 15 training tasks and 30 validation tasks, with the held-out test set reserved for final evaluation. Within each benchmark--backbone setting, all four optimizers share the same $S_0$, training tasks, and interaction environment.

We conduct independent experiments with four LLM backbones: GLM-5.3-Flash~\cite{zeng2026glm}, DeepSeek-V4.1-Flash~\cite{xu2026deepseek}, GPT-6-Luna~\cite{openai2026gpt6luna}, and Qwen3.8-Max~\cite{qwen38}. Each backbone is used consistently for skill generation, optimization, agent execution, and skill composition, including LLM-based baselines. Each optimizer follows its original optimization procedure and early-stopping strategy, subject to a maximum of four optimization rounds. We select the main skill based on the highest pass rate across three evaluations on the training tasks. All methods are evaluated on the same held-out test tasks under consistent execution settings. We report task success rates (\%) and their arithmetic mean across the five benchmarks for overall performance comparison.

\subsection{Main Results}
Table~\ref{tab:main_results} reports the main results across four LLM backbones and five benchmarks. HBT achieves a 74.20\% average success rate across all 20 settings, outperforming the strongest composition baseline on each backbone by 4.51 percentage points on average (up to 90.63\% vs. Trace2Skill's 85.85\% on Qwen3.8-Max) and the best single optimized skill by 6.23 percentage points. While existing composition baselines improve overall averages over individual optimizers, they frequently trail the strongest single optimizer on specific tasks; for instance, Trace2Skill drops to 26.67\% on GPT-6-Luna ScienceWorld (below SkillEvolver's 48.33\%). This indicates that without history-guided evidence analysis and conditional integration, unconditional merging can compromise individual task performance
and induce negative transfer, highlighting HBT's effectiveness in consolidating complementary strengths.

Beyond aggregate gains, HBT generalizes reliably across model scales and task domains. Across the four backbones, HBT surpasses the strongest composition baseline by 5.80, 2.70, 4.75, and 4.78 percentage points on GLM-5.3-Flash, DeepSeek-V4.1-Flash, GPT-6-Luna, and Qwen3.8-Max, respectively. Across the five benchmarks, whereas individual optimizers exhibit narrow task specialization with no single method dominating all domains, HBT consistently captures peak specialized performance across scientific exploration, spreadsheet manipulation, multi-step planning, and tool execution, ranking first or tied first in 17 of 20 settings. This consistent superiority across both backbones and benchmarks confirms the strong generalizability of our framework across diverse agent environments.

\begin{table}[t]
    \centering
    \caption{Ablation results on Spreadsheet, DeepPlanning, and AppWorld using GLM-5.3-Flash. All values are task success rates (\%). Avg. denotes the arithmetic mean across the three benchmarks. Best results are highlighted in bold.}
    \label{tab:ablation}
    \small
    \setlength{\tabcolsep}{2.5pt}
    \begin{tabular}{lcccc}
        \toprule
        Method & Spreadsheet & \shortstack{DeepPlanning} & AppWorld & Avg. \\
        \midrule
        Full            & \textbf{73.21} & \textbf{53.75} & \textbf{76.79} & \textbf{67.92} \\
        \midrule
        w/o History             & 72.50 & 43.75 & 70.83 & 62.36 \\
        Rand.Trigger               & 68.21 & 41.25 & 66.67 & 58.71 \\
        Rand.Boundary              & 72.86 & 46.25 & 63.69 & 60.93 \\
        Positive-only           & \textbf{73.21} & 48.75 & 70.24 & 64.07 \\
        \bottomrule
    \end{tabular}
\end{table}
\begin{table}[t]
    \centering
    \caption{Sensitivity to main skill selection across three benchmarks. Each w/ row reports HBT performance when the skill produced by the optimizer is selected as the main skill. HBT selects the main skill based on training-set performance for each benchmark.}
    \label{tab:main_skill}
    \small
    \setlength{\tabcolsep}{5pt}
    \begin{tabular}{lcccc}
        \toprule
        Main Skill & Spreadsheet & DeepPlanning & AppWorld \\
        \midrule
        HBT              & 73.21 & 53.75 & 76.79  \\
        w/ SkillOpt       & 74.64 & 52.50 & 77.98  \\
        w/ SkillHone      & 72.86 & 53.75 & 76.79  \\
        w/ EvoSkill       & 73.21 & 50.00 & 76.19  \\
        w/ SkillEvolver   & 73.21 & 48.75 & 74.40  \\
        \bottomrule
    \end{tabular}
\end{table}

\section{Ablation and Further Analysis}

\subsection{Component Ablations}

We conduct four component-level ablations on Spreadsheet, DeepPlanning, and AppWorld using GLM-5.3-Flash. The results are summarized in Table~\ref{tab:ablation}. For the randomized variants, Rand.Trigger and Rand.Boundary, we repeat the experiments with multiple random seeds and report the average performance across independent runs.

\noindent\textbf{Historical Evidence and Boundary Localization.}
The w/o History variant infers behavioral applicability conditions solely from optimized skills' execution trajectories, without accessing optimization histories, dropping 5.56 percentage points below HBT to 62.36\%. This highlights the value of historical evidence for reliable condition inference. Rand.Boundary retains trade-off analysis and condition inference, but randomly samples a revision round among those with high behavior occurrence rather than localizing the transition boundary. Its average success rate drops further to 60.93\%, notably on AppWorld, as high-frequency plateaus provide less informative evidence about behavioral changes, underscoring the value of our shift-based localization in identifying informative revision boundaries. Additional experiments in the Supplementary Material examine the impact of execution stochasticity on boundary localization, further confirming its robustness.

\noindent\textbf{Condition Inference.}
Rand.Trigger retains HBT's candidate behaviors and activation frequency but triggers them randomly across test tasks, dropping average success rate to 58.71\%. This confirms the advantage of our state-aware condition inference in preventing indiscriminate activation and routing behaviors to matching contexts. Positive-only infers conditions solely from beneficial cases (FAIL-to-PASS) while excluding detrimental cases (PASS-to-FAIL), lowering average success rate to 64.07\%, notably on DeepPlanning and AppWorld. Without negative counterexamples, inferred conditions become overly permissive, underscoring the strength of our contrastive inference in delineating applicability boundaries. Experiments in the Supplementary Material verify that condition-guided behaviors activate in alignment with their inferred applicability conditions.

\begin{table}[t]
    \centering
    \caption{Behavioral evidence evaluation across three benchmarks. Label agreement (\%) measures the proportion of predictions matching ablation-derived reference labels. Avg. denotes the mean across benchmarks.}

    \label{tab:evidence}
    \small
    \setlength{\tabcolsep}{5pt}
    \begin{tabular}{@{}lcccc@{}}
        \toprule
        Method & Spreadsheet & DeepPlanning & AppWorld & Avg. \\
        \midrule
        Trajectory  & 61.34 & 55.83 & 67.55 & 61.57 \\
        Result-only & 68.72 & 56.08 & 60.65 & 61.82 \\
        HBT        & \textbf{81.30} & \textbf{74.15} & \textbf{76.18} & \textbf{77.21} \\
        \bottomrule
    \end{tabular}
\end{table}

\subsection{Integration Placement}
Our default strategy appends the inferred condition-behavior pairs to the end of the main skill. To examine whether inline insertion improves composition, we evaluate LLM Placement, which prompts an LLM to insert each pair into a suitable position within the main skill. While this variant slightly outperforms HBT on Spreadsheet (73.93\% vs. 73.21\%), it degrades performance on DeepPlanning (47.50\% vs. 53.75\%) and AppWorld (67.86\% vs. 76.79\%), lowering average success rate from 67.92\% to 63.10\%. Modifying the internal structure of the main skill risks fragmenting its coherent workflow and introducing unintended instruction interference. By contrast, appending explicit conditional rules preserves the structural integrity of the main skill while relying on pre-decision condition matching to trigger specialized behaviors at the appropriate execution steps.

\begin{figure}[t]
    \centering
    \includegraphics[width=\columnwidth]{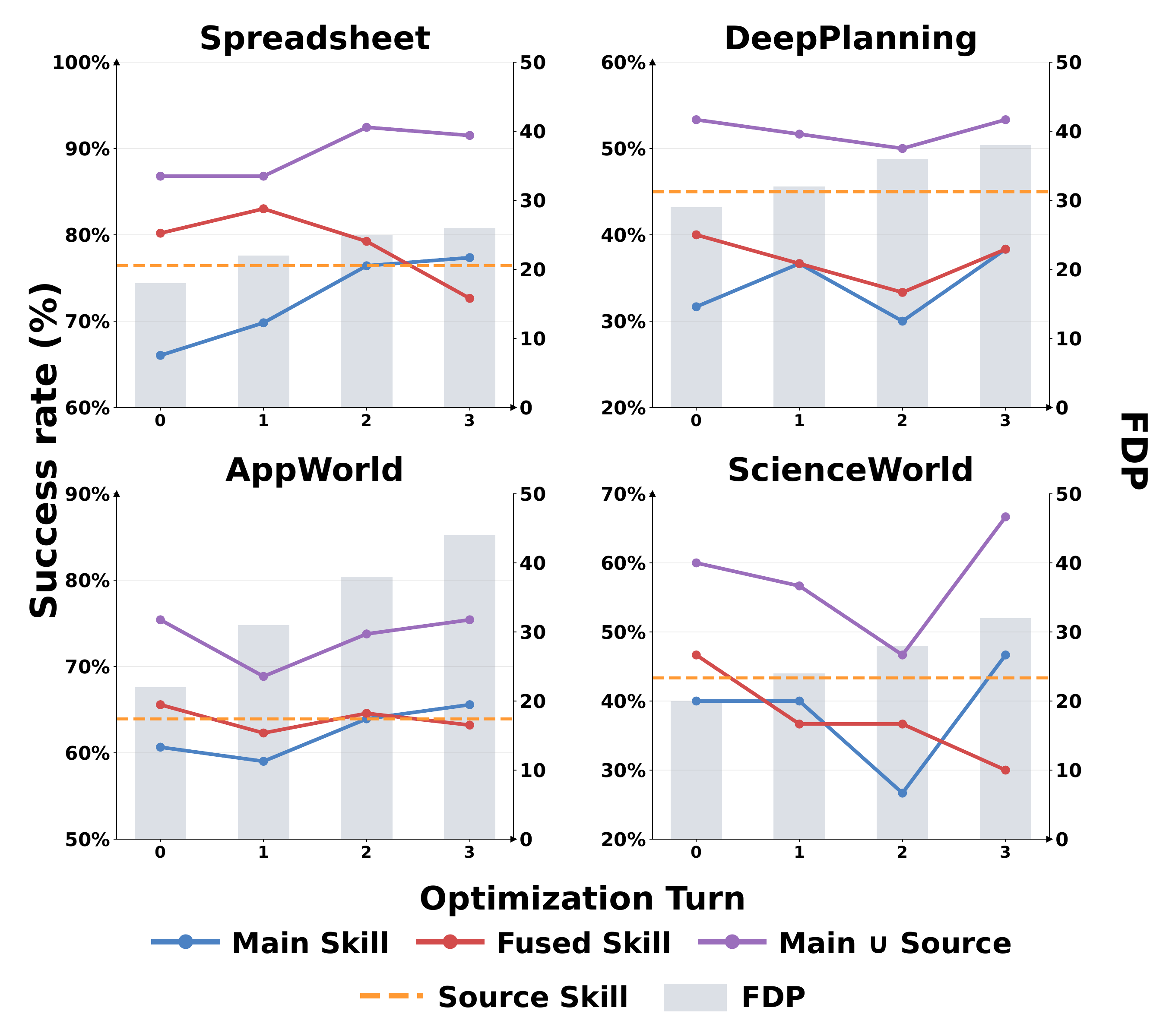}
    \caption{Turn-wise skill fusion performance and Fusion Decision Points (FDPs) across four benchmarks. Curves (left axis) report task success rates for the intermediate main skill (Main Skill), the fixed source skill (Source Skill), their direct fusion (Fused skill), and their combined task coverage (Main $\cup$ Source). Gray bars (right axis) show the number of FDPs at each optimization turn.}
    \label{fig:motivation}
\end{figure}

\subsection{Sensitivity to Main Skill Selection}

Our default setting selects the main skill as the optimized skill with the highest success rate on the training tasks. We adopt this criterion to avoid an additional cross-optimizer selection stage, which would require uniformly re-evaluating all four optimized skills on a validation set. However, training performance may not always identify the best-performing candidate on unseen tasks. To examine the sensitivity of HBT to this choice, we evaluate our composition procedure by designating each of the four optimized skills as the main skill in turn.

As shown in Table~\ref{tab:main_skill}, the training-based choice does not always match the top-performing candidate on the test set. Nevertheless, composing from the training-selected main skill achieves an average success rate of 67.92\%, within 0.87 percentage points of using the per-benchmark test-best main skill (68.79\%). This narrow gap indicates that although minor discrepancies exist between training and test rankings, our training-based selection criterion is robust and incurs negligible performance loss.

\subsection{Behavioral Evidence Evaluation}

To evaluate how well different evidence strategies distinguish beneficial from detrimental behaviors, we derive reference labels through behavior-level ablations on Spreadsheet, DeepPlanning, and AppWorld. For each candidate behavior, we remove it from the post-boundary skill while keeping other instructions unchanged, and re-evaluate the task. PASS-to-FAIL and FAIL-to-PASS transitions indicate beneficial and detrimental behaviors, respectively; unchanged outcomes are excluded. We compare three strategies: Result-only, using cross-revision outcome changes; Trajectory, examining isolated post-boundary trajectories; and HBT, contrasting stage-aligned paired trade-off trajectories.

As shown in Table~\ref{tab:evidence}, HBT achieves the highest agreement with reference labels across all three benchmarks, averaging 77.21\%, outperforming Result-only and Trajectory by 15.39 and 15.64 percentage points, respectively. These results suggest that paired trajectory comparison provides more informative evidence for distinguishing beneficial and detrimental behaviors than outcome changes or isolated trajectories alone.

\begin{table}[t]
    \centering
    \caption{Condition-wise rollback analysis across three benchmarks. Restored pre-decision states ($N$) are grouped by whether inferred applicability conditions hold. Gain and Loss denote FAIL-to-PASS and PASS-to-FAIL rates (\%) under unconditional behavior transfer relative to the main skill; Net is calculated as Gain minus Loss.}
    \label{tab:condition_analysis}
    \small
    \setlength{\tabcolsep}{4.2pt}
    \begin{tabular}{lrrrrrr}
        \toprule
        & \multicolumn{2}{c}{Spreadsheet} & \multicolumn{2}{c}{DeepPlanning} & \multicolumn{2}{c}{AppWorld} \\
        \cmidrule(lr){2-3} \cmidrule(lr){4-5} \cmidrule(lr){6-7}
        Metric & True & False & True & False & True & False \\
        \midrule
        $N$       & 61   & 191   & 34   & 46    & 57   & 105   \\
        Gain (\%) & 14.8 & 2.6   & 17.6 & 8.7   & 14.0 & 9.5   \\
        Loss (\%) & 8.2  & 8.9   & 14.7 & 21.7  & 7.0  & 29.5  \\
        \midrule
        Net (pp)  & $\mathbf{+6.6}$ & $-6.3$ & $\mathbf{+2.9}$ & $-13.0$ & $\mathbf{+7.0}$ & $-20.0$ \\
        \bottomrule
    \end{tabular}
\end{table}

\subsection{Behavioral Interference in Direct Fusion}

To examine the challenges of direct skill fusion, we fuse each intermediate skill along the SkillEvolver trajectory with a fixed optimized skill from another optimizer (Source Skill) across four benchmarks. As shown in Fig.~\ref{fig:motivation}, direct fusion yields diminishing gains and even performance regressions at later optimization turns, while consistently trailing the combined task coverage of the two skills (Main $\cup$ Source). Meanwhile, the gray bars track Fusion Decision Points (FDPs), defined as distinct decision points where the two skills prescribe differing actions for the same execution stage. Across all four benchmarks, FDPs increase steadily over optimization turns, indicating escalating coordination complexity as skills evolve during joint execution.

As skills specialize, their growing conflicts make direct fusion harder. HBT addresses this interference through state-conditioned activation.

\subsection{Analysis of Applicability Conditions}

To assess the importance of conditional activation, we evaluate a variant that transfers candidate behaviors unconditionally across execution states. Across all five benchmarks, this variant reduces the average success rate by 12.98 percentage points, from 68.25\% to 55.27\%. To further examine whether inferred conditions distinguish beneficial from detrimental transfers, we conduct a paired rollback analysis on Spreadsheet, DeepPlanning, and AppWorld. Specifically, we restore test execution to the pre-decision state of each candidate behavior and compare two continuations from the same state: running the main skill alone versus applying transferred behaviors unconditionally. We group these states according to whether the inferred conditions hold (True vs. False) and report Gain (FAIL-to-PASS rate), Loss (PASS-to-FAIL rate), and Net (Gain minus Loss).

As shown in Table~\ref{tab:condition_analysis}, unconditional transfer yields positive Net gains across all three benchmarks in the True group, whereas losses outweigh gains in the False group, resulting in consistently negative Net scores. These results further demonstrate that the inferred conditions distinguish states where transferred behaviors are beneficial from those where they cause interference, supporting the effectiveness of state-conditioned activation across different pre-decision states.

\begin{table}[t]
    \centering
    \caption{Accuracy (\%) on V*Bench across attribute recognition, spatial reasoning, and overall evaluation for two VLM backbones. Overall denotes task-level accuracy across the full benchmark rather than the unweighted arithmetic mean of the two sub-categories. Best results per backbone are in bold.}
    \label{tab:vstar}
    \small
    \setlength{\tabcolsep}{12pt}
    \begin{tabular}{lccc}
        \toprule
        Method & Attribute & Spatial & Overall \\
        \midrule
        \multicolumn{4}{l}{\textit{GLM-5.3-Flash}} \\
        \midrule
        Init Skill    & 81.44 & 70.31 & 77.02 \\
        SkillOpt      & 86.60 & 75.00 & 81.99 \\
        SkillHone     & 86.60 & 70.31 & 80.12 \\
        EvoSkill      & 81.44 & 68.75 & 76.40 \\
        SkillEvolver  & 85.57 & 75.00 & 81.37 \\
        \textbf{HBT} & \textbf{89.69} & \textbf{76.56} & \textbf{84.47} \\
        \midrule
        \multicolumn{4}{l}{\textit{DeepSeek-V4.1-Flash}} \\
        \midrule
        Init Skill    & 69.07 & 81.25 & 73.91 \\
        SkillOpt      & 77.32 & 85.94 & 80.75 \\
        SkillHone     & 76.29 & 81.25 & 78.26 \\
        EvoSkill      & 73.20 & 82.81 & 77.02 \\
        SkillEvolver  & 74.23 & 90.63 & 80.75 \\
        \textbf{HBT} & \textbf{78.35} & \textbf{92.19} & \textbf{83.85} \\
        \bottomrule
    \end{tabular}
\end{table}

\subsection{Generalization to Visual Tasks}

Following prior work on visual skill self-evolution~\cite{wang2026visualevolve}, we evaluate HBT on V*Bench~\cite{wu2024v}, which tests attribute recognition and spatial reasoning in high-resolution images. Against the initial skill and skills from four optimizers, HBT achieves 84.47\% on GLM-5.3-Flash and 83.85\% on DeepSeek-V4.1-Flash (84.16\% on average), outperforming the best single optimized skill by 2.48 and 3.10 percentage points, respectively (Table~\ref{tab:vstar}).

\section{Conclusion}
HBT uses optimization histories to assess when behavioral changes help or hurt execution. It extracts behavioral differences, localizes revision boundaries, and contrasts beneficial and detrimental transitions to infer applicability conditions. By integrating condition-behavior pairs into a main skill, HBT mitigates context-insensitive behavior transfer while preserving a coherent skill structure. Experiments across five benchmarks and four skill optimizers demonstrate consistent improvements over individual skills and composition baselines, with V*Bench results extending these findings to visual tasks.

\clearpage
 {
     \small
     \bibliographystyle{ieeenat_fullname}
     \bibliography{main}
 }

\clearpage
\setcounter{page}{1}
\maketitlesupplementary

\section{Method}
\label{sec:supp_method}

We provide additional details on the three stages of HBT, focusing on behavior extraction, history tracing, and condition inference.

\subsection{Behavioral Difference Extraction}

\paragraph{Task-level comparison.}
We compare the execution trajectories of the main skill and each source skill on the same training tasks. Source trajectories are considered whenever they succeed, regardless of the main skill's outcome. An LLM identifies behaviors exhibited by the source but absent from the main trajectory, grounding each difference in actual actions and environmental feedback. The prompt used for this step is provided in Table~\ref{tab:prompt_delta}.

\paragraph{Behavior consolidation.}
Task-level differences are consolidated into reusable behaviors based on their underlying decision mechanisms rather than superficial similarities in tools or objects. Instance-specific details are abstracted into functional roles, while essential operations, action ordering, and constraints are preserved. Each candidate retains its supporting execution evidence and source-skill associations, potentially including multiple sources.

\subsection{Behavioral Boundary Localization}

For each candidate behavior $b$, we trace its occurrence across the optimization histories of its source skills. For behaviors originating from $S_0$, we instead examine the main skill's optimization history. An LLM determines behavioral occurrence from recorded actions and observations, and we compute occurrence rates and identify the largest absolute change between adjacent versions following Eqs.~(3) and (4) of the main paper.

For behaviors associated with multiple sources, boundaries are localized independently in each optimization history. The resulting evidence is subsequently aggregated without mixing revision sequences from different optimizers.

\begin{table}[t]
    \centering
    \caption{Selected instructions from the prompt for task-level delta extraction (Stage 1).}
    \label{tab:prompt_delta}
    \small
    \begin{tabularx}{\columnwidth}{@{}>{\raggedright\arraybackslash}X@{}}
        \toprule
        \textit{Prompt for Task-level Delta Extraction} \\

        \textbf{Task-level Behavior Extraction}

        Compare the supplied Main and Other executions of the SAME TASK using actual actions, pre-decision states, environment feedback, and recorded outcomes. Identify meaningful local decision differences; cross-task policy synthesis is a separate step.

        \par\smallskip
        Ground each difference in original execution evidence and cite exact step references from BOTH traces. Thought is not execution; a final PASS does not establish utility, and NOT\_OBSERVED does not establish absence. Do not invent operations or abstract principles without operational meaning. Keep coherent action bundles together and independent differences separate.

        \par\smallskip
        Consider both Main FAIL/Other PASS and distinct effective PASS/PASS paths. Do not extract a behavior already effectively executed by Main. Preserve the supplied source provenance, treating S0 as a special source only when explicitly designated. Do not infer Conditions or localize historical boundaries at this step.
        \\
        \bottomrule
    \end{tabularx}
\end{table}

\begin{table}[t]
    \centering
    \caption{Selected instructions from the prompt for historical trade-off analysis.}
    \label{tab:prompt_tradeoff}
    \small
    \begin{tabularx}{\columnwidth}{@{}>{\raggedright\arraybackslash}X@{}}
        \toprule
        \textit{Prompt for Historical Trade-off Analysis} \\

        \textbf{Historical Boundary and Source Analysis}

        Analyze the Stage 1 Transformation at the Stage 2-selected adjacent historical boundary using the supplied Stage 3 execution structures. Compare chronological parent and child versions. The boundary represents a change in behavior occurrence, not a causal introducing edit.

        \par\smallskip
        Analyze each source endpoint independently using only its own historical lineage and execution evidence. For optimizer-sourced candidates, trace history backward from the source endpoint; for explicitly S0-sourced candidates, trace forward along the Main ancestry. Aggregate source-specific findings only after independent analyses.

        \par\smallskip
        \textbf{Outcome Transition and Local Utility Assessment}

        Focus exclusively on same-task outcome reversals: FAIL-to-PASS and PASS-to-FAIL. Select the corresponding executions from the adjacent parent and child versions. Do not include PASS/PASS or FAIL/FAIL cases. Align comparisons using pre-decision observable states and actual actions, rather than stage names or positions alone. Episodes from the same task are not necessarily identical microstates or strict counterfactuals.

        \par\smallskip
        Trace BEFORE state $\rightarrow$ actual target behavior $\rightarrow$ environmental feedback/local progress $\rightarrow$ final outcome. For FAIL-to-PASS, examine whether the target behavior was executed in the successful episode and whether its observed effects plausibly contributed to success. For PASS-to-FAIL, examine whether the target behavior was executed in the failed episode and whether there is evidence linking it to the regression. Distinguish failures occurring before, during, or after target behavior execution.

        \par\smallskip
        Separate observed facts from plausible interpretations and alternative explanations, including concurrent modifications and execution randomness. Classify local utility as SUPPORTING, OPPOSING, or UNRESOLVED. Supporting or opposing judgments require a chronological chain of pre-execution state, actual target action, and subsequent feedback from the SAME episode. If the behavior was not executed, or the evidence is insufficient or confounded, use UNRESOLVED. Neither final PASS nor behavior occurrence alone establishes utility. Do not infer Conditions or claim causal certainty.
        \\
        \bottomrule
    \end{tabularx}
\end{table}

\subsection{Contrastive Condition Inference}

\paragraph{Trade-off analysis.}
At each localized boundary, we identify tasks exhibiting FAIL-to-PASS or PASS-to-FAIL transitions. Since paired trajectories may differ in length and execution order, an LLM organizes them into functional decision stages and aligns the stages relevant to the candidate behavior. Using the successful trajectory as a reference for subgoal completion, we compare pre-execution states, actual actions, and environmental feedback to identify local progress or regression associated with the behavior. Our goal is to assess the plausible local utility of candidate behaviors from execution evidence, rather than establish strict causal effects. Cases with insufficient or confounded evidence are classified as unresolved rather than treated as supporting or opposing evidence. The prompt used for this analysis is provided in Table~\ref{tab:prompt_tradeoff}.

\paragraph{Condition inference.}
We collect pre-decision states from retained beneficial and detrimental cases as positive and negative evidence, respectively. Evidence from multiple source histories is aggregated for each behavior. An LLM contrasts these states to infer applicability conditions that depend only on information observable before the relevant decision.

\paragraph{Verification and integration.}
An additional verification step checks behavioral grounding, consistency between conditions and historical evidence, and potential conflicts with the main skill. This reduces inconsistencies introduced by multi-stage LLM analysis. Validated condition-behavior pairs are appended to the main skill, instructing the agent to apply transferred behaviors only when their conditions hold.

\section{Benchmark Descriptions}
\label{sec:supp_benchmarks}

\subsection{ScienceWorld}

ScienceWorld is a text-based interactive environment requiring scientific reasoning and multi-step experimentation. We use the Genetics domain, where agents investigate Mendelian inheritance through observations and experiments involving plant traits. Tasks require coordinated actions and reasoning over environmental feedback.

\subsection{SpreadsheetBench}

SpreadsheetBench evaluates spreadsheet manipulation through real-world user requests involving data processing, formulas, and structural modifications. We use the Verified subset, which provides expert-annotated tasks and reliable evaluation references. Successful execution requires translating natural-language instructions into accurate spreadsheet operations.

\subsection{DeepPlanning}

DeepPlanning evaluates long-horizon planning under verifiable constraints. We adopt the Shopping domain, where agents search for products, compare alternatives, and construct purchasing plans satisfying item requirements, discounts, and budget constraints. These tasks require coordinating local decisions with global planning objectives.

\subsection{AppWorld}

AppWorld evaluates agents on multi-application tasks involving API interactions and persistent state changes. We use the official Test-Normal split. Tasks require identifying appropriate APIs, coordinating tool calls, and modifying application states to satisfy natural-language requests.

\subsection{$\tau^3$-Bench}

$\tau^3$-Bench evaluates conversational agents in policy-constrained customer-service environments. We use the Retail and Airline domains, covering operations such as order modifications, returns, flight changes, and cancellations. Agents must coordinate user interactions, policy compliance, and tool execution.

\subsection{V*Bench}

V*Bench evaluates fine-grained visual understanding in high-resolution images through Attribute Recognition and Spatial Relationship Reasoning. The benchmark contains 191 questions requiring precise localization of visual details and reasoning about object properties or spatial relationships. We use it to assess the transferability of conditional skill composition to visual reasoning.

\section{Implementation Details}
\label{sec:supp_implementation}

\subsection{Dataset Splits and Initial Skills}

For each primary benchmark, we use 15 training tasks and 30 validation tasks. For V*Bench, we use 15 training and 15 validation questions, leaving 161 questions for testing. We adopt official test splits when available; otherwise, the remaining tasks form the test set. All methods share identical splits.

HBT uses only training-task trajectories for skill composition. Validation data is reserved for optimizers requiring validation-based selection or early stopping, while test data is used exclusively for final evaluation.

For each benchmark and backbone, we generate a shared initial skill $S_0$ from benchmark and training-task descriptions, excluding solutions, trajectories, and feedback.

\subsection{Skill Optimization and Model Configuration}

We employ SkillOpt, SkillHone, EvoSkill, and SkillEvolver, each following its original optimization and early-stopping procedure with at most four rounds. Intermediate skill versions and execution histories are retained. The main skill is selected by aggregate pass rate across three evaluations on the fixed training tasks.

Experiments use GLM-5.3-Flash, DeepSeek-V4.1-Flash, GPT-6-Luna, and Qwen3.8-Max. Within each setting, the same backbone is used for skill generation, optimization, execution, and composition.

\subsection{Composition Baselines}

\paragraph{Trace2Skill.}
We adapt Trace2Skill by pooling the training-task trajectories of all four optimized skills and applying its trajectory-based skill construction procedure to produce a unified skill. This preserves its experience-driven synthesis mechanism while extending its inputs to multiple optimized skills, without using intermediate optimization histories.

\paragraph{Rewrite--Critique--Revise.}
Our RCR baseline uses three LLM calls: rewriting the four optimized skills into a unified draft, critiquing instruction conflicts and redundancies, and revising the draft into the final skill. It uses endpoint skill instructions without historical trajectories.

\paragraph{Other approaches.}
We also considered SkillComposer, SkillAligner, and SkillCAT. However, complete implementations and required artifacts were not publicly available for faithful reproduction in our setting. Moreover, their learned composition, runtime adaptation, or topology-aware execution mechanisms cannot be directly transferred to our offline composition protocol without substantial modifications. We therefore compare against the reproducible Trace2Skill adaptation and RCR baseline.

\subsection{Evaluation Protocol}

All methods are evaluated on the same held-out test tasks under consistent execution settings. We report task success rates and their arithmetic mean across the five primary benchmarks. For V*Bench, we evaluate GLM-5.3-Flash and DeepSeek-V4.1-Flash, reporting Attribute Recognition, Spatial Relationship Reasoning, and overall accuracy on the 161 test questions.

\subsection{Computational Overhead}

HBT introduces additional computational overhead due to
its reliance on historical execution analysis. In particular,
processing intermediate skill versions and their associated
trajectories requires multiple LLM-based operations, which can
be substantially more expensive than direct composition from
endpoint skills. We recognize this as a practical trade-off
of our approach rather than claiming computational efficiency
over history-free composition methods.

Nevertheless, optimization histories can be retained and
reused once generated, avoiding repeated optimization runs
when constructing or updating composed skills. Thus, the
additional cost primarily reflects the offline analysis of
historical evidence, which provides access to behavioral
transitions and execution-level evidence unavailable to
methods relying solely on endpoint skills.

\begin{table}[t]
    \centering
    \caption{Additional experimental results using GLM-5.3. SciW., Spread., DeepP., AppW., and $\tau^3$B. denote ScienceWorld, SpreadsheetBench, DeepPlanning, AppWorld, and $\tau^3$-Bench, respectively. All values are task success rates (\%), with the best results highlighted in bold.}
    \label{tab:glm_results}
    \small
    \setlength{\tabcolsep}{2.5pt}
    \renewcommand{\arraystretch}{1.12}
    \begin{tabular*}{\columnwidth}
    {@{\extracolsep{\fill}}lrrrrrr@{}}
        \toprule
        Method & SciW. & Spread. & DeepP. & AppW. & $\tau^3$B. & Avg. \\
        \midrule
        $S_0$         & 35.00 & 65.71 & 36.25 & 61.31 & 80.60 & 55.77 \\
        SkillOpt      & 27.50 & 74.64 & 40.00 & 71.43 & 82.84 & 59.28 \\
        SkillHone     & 35.00 & 69.29 & 45.00 & 56.55 & 85.82 & 58.33 \\
        EvoSkill      & 45.00 & 73.21 & 37.50 & 63.10 & 85.07 & 60.78 \\
        SkillEvolver  & 45.00 & 74.29 & 48.75 & 79.17 & 80.60 & 65.56 \\
        \midrule
        Trace2Skill   & 48.33 & 73.57 & \textbf{52.50} & 74.40 & 89.55 & 67.67 \\
        RCR           & 41.67 & 71.43 & 41.25 & 77.38 & 90.30 & 64.41 \\
        \midrule
        HBT     & \textbf{55.00} & \textbf{75.36} & 48.75
                      & \textbf{80.36} & \textbf{91.04} & \textbf{70.10} \\
        \bottomrule
    \end{tabular*}
\end{table}

\section{Extended Experiment}
\subsection{Additional Results with GLM-5.3}
\label{sec:glm_results}

To further evaluate the effectiveness of HBT across different LLM backbones, we report additional experimental results using GLM-5.3, which are not included in the main paper. As shown in Table~\ref{tab:glm_results}, HBT achieves the highest average success rate of 70.10\% across the five benchmarks, outperforming all compared optimization and composition methods. These results further demonstrate the effectiveness of history-guided conditional skill composition across different LLM backbones.

\begin{table}[t]
    \centering
    \caption{Stability evaluation of HBT using GLM-5.3-Flash over three independent runs. Results are reported as mean $\pm$ standard deviation (\%).}
    \label{tab:stability}
    \small
    \setlength{\tabcolsep}{6pt}
    \renewcommand{\arraystretch}{1.15}
    \begin{tabular}{lc}
        \toprule
        Benchmark & Mean $\pm$ Std. (\%) \\
        \midrule
        ScienceWorld      & 51.39 $\pm$ 1.27 \\
        SpreadsheetBench  & 72.86 $\pm$ 0.94 \\
        DeepPlanning      & 53.33 $\pm$ 0.72 \\
        AppWorld          & 76.59 $\pm$ 0.91 \\
        $\tau^3$-Bench    & 85.07 $\pm$ 0.75 \\
        \midrule
        Average           & 67.85 $\pm$ 0.35 \\
        \bottomrule
    \end{tabular}
\end{table}

\subsection{Comparison with Alternative Skill Composition Strategies}
\label{sec:composition_baselines}

We further compare HBT with two alternative strategies for leveraging optimized skills, using GLM-5.3-Flash across five benchmarks. The first baseline, \textit{Concat.}, directly concatenates the optimized skills from different optimizers into a single skill without further analysis or integration. The second baseline, \textit{LLM Select.}, uses an LLM to select a skill from the candidate optimized skills based on the current task description. Specifically, the LLM receives the task description and candidate skills, and selects the skill it considers most relevant to the task, which is then used for execution.

As shown in Table~\ref{tab:skill_selection}, LLM-based selection consistently outperforms direct concatenation, indicating that task-aware skill selection is more effective than indiscriminately combining multiple skills. Nevertheless, HBT achieves the best performance across all five benchmarks, with an average success rate of 68.25\%, compared with 63.63\% for LLM Select. These results suggest that identifying and transferring useful behaviors from optimization histories, together with their execution conditions, provides additional benefits beyond selecting existing skills based solely on task descriptions.

\begin{table}[t]
    \centering
    \caption{Task success rates (\%) of different skill construction and selection strategies across five benchmarks using GLM-5.3-Flash. Avg. denotes the unweighted mean across benchmarks. Benchmark abbreviations follow Table~\ref{tab:glm_results}.}
    \label{tab:skill_selection}
    \small
    \setlength{\tabcolsep}{3pt}
    \begin{tabular*}{\columnwidth}
    {@{\extracolsep{\fill}}lcccccc@{}}
        \toprule
        Method & SciW. & SB. & DeepP. & AppW. & $\tau^3$B. & Avg. \\
        \midrule
        Concat.     & 33.33 & 65.36 & 25.00 & 61.31 & 83.58 & 53.72 \\
        LLM Select. & 48.33 & 71.79 & 43.75 & 71.43 & 82.84 & 63.63 \\
        \midrule
        HBT & \textbf{51.67} & \textbf{73.21} & \textbf{53.75} & \textbf{76.79} & \textbf{85.82} & \textbf{68.25} \\
        \bottomrule
    \end{tabular*}
\end{table}

\subsection{Stability Analysis}
\label{sec:stability}

HBT involves multiple LLM-based operations, including behavioral difference extraction, behavior consolidation, historical trajectory analysis, decision-stage alignment, condition inference, and evidence verification. Since these operations may introduce stochasticity at different stages, it is important to examine whether such variations propagate to the final composed skill and affect its performance.

To this end, we independently run the complete HBT pipeline three times using GLM-5.3-Flash under the same experimental settings and evaluate the resulting skills on the five benchmarks. We report the mean and standard deviation of task success rates across the three runs, as summarized in Table~\ref{tab:stability}.

The results show relatively small performance variations across repeated runs. In particular, the standard deviations remain below 1.3 percentage points on all five benchmarks, while the overall average success rate is $67.85\% \pm 0.35\%$. Despite the multiple stages of LLM-based reasoning involved in the pipeline, the resulting performance remains consistent across independent runs. This suggests that HBT is reasonably robust to the stochasticity of LLM-based processing, and that its history-guided behavior extraction and conditional composition process can produce stable improvements in practice.

\begin{figure*}[t]
    \centering
    \includegraphics[width=\textwidth]{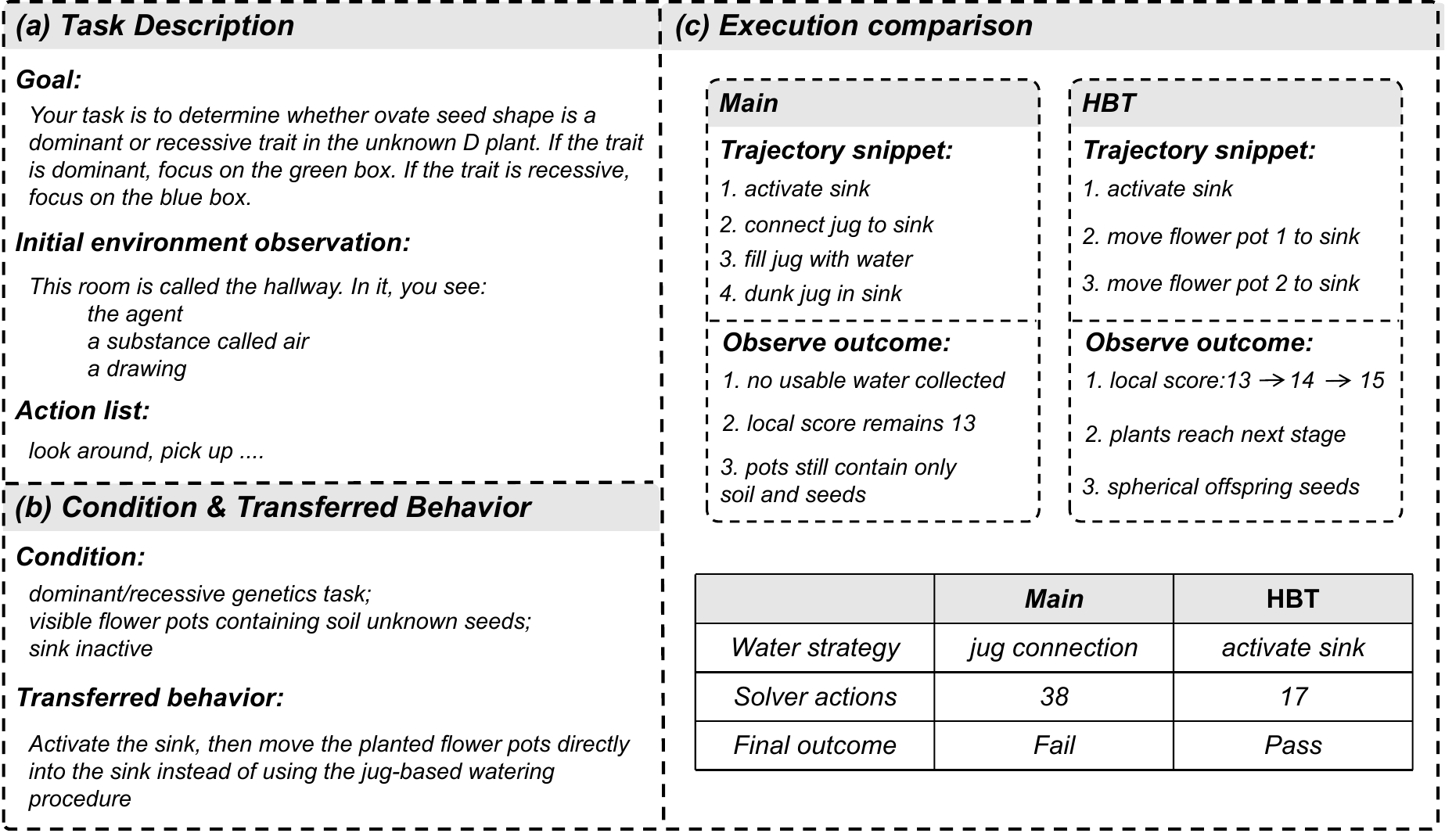}
    \caption{Case study on conditional watering strategy transfer in ScienceWorld Genetics (task v385). The figure illustrates the task description, the inferred initiation condition and transferred behavior, and the execution comparison between Main and HBT.}
    \label{fig:case_study}
\end{figure*}

\subsection{Reliability of Behavioral Boundary Localization}
\label{sec:boundary_reliability}

A potential concern with our behavioral boundary localization is that agent executions are inherently stochastic. Consequently, the identified boundary, determined by the largest change in behavioral occurrence rates between adjacent skill versions, might be sensitive to execution randomness. To examine its reliability, we conduct an additional experiment using GLM-5.3-Flash. Specifically, for selected optimizer histories and their corresponding training tasks where HBT has identified behavioral boundaries, we independently execute each intermediate skill version three additional times. We then reapply our behavioral occurrence analysis and boundary localization procedure and compare the re-identified boundaries with the original ones.

The results show that all re-identified boundaries exactly match the originally identified boundaries, achieving 100\% localization consistency. Furthermore, our statistics and observations across all collected data reveal a clear distinction in behavioral occurrence rates: consistently exhibited behaviors typically have occurrence rates above 60\%, whereas occasional occurrences generally remain below 20\%. This suggests that skill instructions systematically influence the likelihood of executing particular behaviors, making persistent behavioral tendencies distinguishable from sporadic occurrences.

Together, these findings indicate that the largest change in behavioral occurrence rate captures relatively stable transitions in behavior induced by skill revisions, rather than isolated stochastic events. They therefore support the reliability of our approach to identifying informative historical boundaries.

\begin{table}[t]
    \centering
    \caption{Condition-behavior triggering accuracy (\%) on held-out test trajectories across five benchmarks and four LLM backbones. A behavior is correctly triggered when its condition is satisfied and correctly withheld when its condition is not satisfied. Avg. denotes the unweighted mean across benchmarks. Benchmark abbreviations follow Table~\ref{tab:glm_results}.}
    \label{tab:conditional_execution}
    \small
    \setlength{\tabcolsep}{3pt}
    \renewcommand{\arraystretch}{1.2}
    \begin{tabular*}{\columnwidth}
    {@{\extracolsep{\fill}}c*{6}{c}@{}}
        \toprule
        Backbone & SciW. & Spread. & DeepP. & AppW. & $\tau^3$B. & Avg. \\
        \midrule
        \multirow[c]{2}{*}{\shortstack[c]{GLM\\5.3-Flash}}
        & \multirow[c]{2}{*}{95.12}
        & \multirow[c]{2}{*}{93.96}
        & \multirow[c]{2}{*}{86.81}
        & \multirow[c]{2}{*}{93.11}
        & \multirow[c]{2}{*}{85.11}
        & \multirow[c]{2}{*}{90.82} \\
        & & & & & & \\
        \addlinespace[3pt]

        \multirow[c]{2}{*}{\shortstack[c]{DeepSeek\\V4.1-Flash}}
        & \multirow[c]{2}{*}{86.73}
        & \multirow[c]{2}{*}{86.85}
        & \multirow[c]{2}{*}{83.93}
        & \multirow[c]{2}{*}{96.95}
        & \multirow[c]{2}{*}{92.07}
        & \multirow[c]{2}{*}{89.31} \\
        & & & & & & \\
        \addlinespace[3pt]

        \multirow[c]{2}{*}{\shortstack[c]{GPT\\6-Luna}}
        & \multirow[c]{2}{*}{93.14}
        & \multirow[c]{2}{*}{94.59}
        & \multirow[c]{2}{*}{90.72}
        & \multirow[c]{2}{*}{100.00}
        & \multirow[c]{2}{*}{87.57}
        & \multirow[c]{2}{*}{93.20} \\
        & & & & & & \\
        \addlinespace[3pt]

        \multirow[c]{2}{*}{\shortstack[c]{Qwen\\3.8-Max}}
        & \multirow[c]{2}{*}{91.04}
        & \multirow[c]{2}{*}{90.59}
        & \multirow[c]{2}{*}{92.37}
        & \multirow[c]{2}{*}{97.23}
        & \multirow[c]{2}{*}{91.55}
        & \multirow[c]{2}{*}{92.56} \\
        & & & & & & \\
        \bottomrule
    \end{tabular*}
\end{table}

\subsection{Condition-Behavior Triggering Accuracy}
\label{sec:conditional_execution}

To evaluate whether the inferred conditions correctly regulate behavior execution, we analyze the execution traces on the held-out test sets across five benchmarks and four LLM backbones. For each condition-behavior pair, we examine instances where the condition is either satisfied or unsatisfied and check whether the corresponding behavior is executed. An instance is considered correctly handled if the behavior is executed when its condition is satisfied and withheld when the condition is not satisfied. We define the condition-behavior triggering accuracy (CBTA) as the proportion of correctly handled instances among all evaluated instances.

As shown in Table~\ref{tab:conditional_execution}, the accuracy remains consistently high across different benchmarks and LLM backbones, with an overall average of 91.47\%. These results indicate that the inferred conditions not only activate their associated behaviors when appropriate but also help prevent their execution when the specified conditions are not met. This demonstrates that HBT produces actionable condition-behavior pairs that regulate behavior execution beyond the training tasks.

\subsection{Case Study}
\label{sec:case_study}

We present a case study from ScienceWorld Genetics (task v385) to illustrate how HBT discovers and transfers useful behaviors from optimization histories. The task requires the agent to determine whether the ovate seed shape of Unknown Plant D is a dominant or recessive trait. To make this determination, the agent must grow the plants and observe their offspring. A green answer box corresponds to a dominant trait, while a blue answer box corresponds to a recessive trait.

\paragraph{Behavior Discovery and Condition Inference.}
By analyzing execution traces from SkillOpt, SkillHone, and SkillEvolver, HBT identifies an alternative watering strategy that bypasses the conventional jug-based procedure: directly moving planted flower pots into an activated sink, avoiding intermediate jug-handling operations.

To determine when this behavior should be applied, HBT examines the execution states and environmental feedback surrounding it in the source traces. In one successful execution, a flower pot containing soil and a planted green unknown B seed is moved into the sink. The subsequent observation confirms that water has been added to the pot, accompanied by an increase in local score from 12 to 13. In a contrasting execution, the seed remains in its jar while the flower pot contains only soil. These observations suggest that the key prerequisite is that the seed has already been planted and the pot still requires watering. The inferred initiation condition also specifies the dominant/recessive genetics task context, linking the behavior to the relevant task family and cultivation stage.

\paragraph{Execution Comparison.}
Figure~\ref{fig:case_study} compares the execution of Main and HBT under the relevant conditions. Before Step 5, both test trajectories contain planted flower pots, and the sink is inactive. Main activates the sink and attempts to connect a jug, fill it with water, and dunk it in the sink. However, these operations fail to provide usable water: the local score remains at 13, and the pots still contain only soil and seeds. Main subsequently selects the green answer box and fails the task after 38 actions.

In contrast, HBT activates the sink and directly moves the two planted flower pots into it. Both movements produce immediate score increases, raising the local score from 13 to 14 and then to 15. The plants subsequently progress to the next growth stage. After further waiting and pollination with a bee, both plants reach the reproduction stage, allowing the agent to observe spherical offspring seeds. Using this evidence, the agent identifies ovate as a recessive trait and selects the blue answer box, successfully completing the task with a score of 100 after 17 actions.

This case illustrates how HBT transfers a strategy from other optimization histories under an appropriate initiation condition, replacing Main's unsuccessful jug-based procedure with direct pot relocation and enabling the subsequent growth, reproduction, and offspring observation required to solve the task.

\end{document}